\pdfoutput=1
\documentclass[review]{elsarticle}
\usepackage{hyperref}
\usepackage{graphicx} 
\usepackage{amsmath}
\usepackage{lineno}
\usepackage{titlesec}
\modulolinenumbers[5]
\usepackage{chngcntr}
\usepackage{marvosym}
\usepackage{url}

\usepackage{multirow}

\begin{document}

\begin{frontmatter}

\title{Phase Contrast Microscopy Cell Population Segmentation: A Survey}

\author{Lin Zhang\\
lzhang22@albany.edu}
\address{University at Albany, SUNY}

\begin{abstract}
Phase contrast microscopy (PCM) has been widely used in the biomedicine, which allows users to observe objectives without staining or killing them.
One important related research is to employ PCM to monitor live cells.
How to segment cell populations in obtained PCM images gains more and more attention as its a critical step for downstream applications, such as cell tracking, cell classification and others.
Many papers have been published to deal with this problem from different perspectives.
In this paper we aim to present a comprehensive review on the development of PCM cell population segmentation.



\end{abstract}

\begin{keyword}
\texttt{phase contrast microscopy (PCM), cell population segmentation}

\end{keyword}

\end{frontmatter}


\section{Introduction}
Phase contrast microscopy was invented by Dutch physicist Frits Zernike \cite{Zernike} using pure optical principles.
In essence, cells are transparent and have different refractive indexes, thus light will be differentiated into two parts when goes through them: one remains the same, and  the other is deviated. PCM manipulates these two parts differently by using optics and phase rings to convert the difference into visible intensity information, i.e. PCM images.
The main advantage of phase contrast microscopy is that cells will be monitored in their natural state without being killed or stained like previously methods, such as fluorescence. 
Therefore, the analysis on the generated PCM images is critical to understand activities of cell. 
This paper focuses on cell segmentation as this is an fundamental process for other downstream applications, such as cell tracking~\cite{kang2008}, cell detection~\cite{Yang2005}, cell classification~\cite{semi-supervised} and others.


Before we present existing methods, we give a formal definition of PCM cell population segmentation problem. Given a PCM image that contains a group of cells, we aim to segment each individual cell in it.
There are three main challenges in PCM cell population segmentation as below: 
\begin{itemize}
\item Densely populated cells: Since many cells cluster in a image,
cell adhesion is inevitable, therefore, the boundary between cells is unclear. 
\item Cell shape deformation: A cell's shape is uncertain due to its activities, such as movements and cell events, such as apoptosis and mitosis, will change the appearance of cells significantly.
\item Artifacts: Due to the imaging principle and imperfection of devices, images often contains artifacts, such as bright halo and shade-off. 
\end{itemize}
Therefore, this problem is different to other cell segmentation problems, such as fluorescence microscopy cell segmentation~\cite{Dzyubachyk01102010,Bergeest2011} and natural image segmentation~\cite{liu2015semantic,7298952}.
In fluorescence microscopy cell images, cells are colored, therefore, they are easy to be separated from their background. 
For natural image segmentation, it has not  artifacts as in PCM cell images, such as bright halo and shade-off. 

In the past years, many algorithms have been proposed to deal with PCM cell population segmentation, however, 
a comprehensive survey on this problem is still missing.
A few papers~\cite{Erik,Gurari} presented partial survey relate to PCM cell segmentation, however, these either too general or too narrow.



The remainder of this paper is structured as follows: A comprehensive analysis of existing cell population segmentation methods is presented in Section 2. In Section 3, we discuss related issues, including databases and performance evaluation. The final conclusion will be given in Section 4.

\section{Analytical methods of phase contrast microscopy cell population segmentation}

In this section, we review papers of phase contrast microscopy cell population segmentation, which are summarized as three categories as below. 
\begin{itemize}
\item 1. Contour-based methods: These methods segment cells from surroundings by using the contour information of cells. 
\item 2. Machine Learning-based methods: These algorithms convert segmentation problem into classification problem. The statistical models are learned to
distinguish cells and non-cell.  
\item 3. Restoration-based methods: In contrast to other methods that treat phase contrast microscopy images as natural images, these rule-based methods aim to 
restore artifact-free images base on the imaging process of phase contrast microscopy. 
Then we can perform cell segmentation on the obtain artifact-free images by thresholding.
\end{itemize}

What needs illustration is that some methods could be assigned into multiple categories and a discussion on this will be given this at the end of this section. Next, we'll analyze the motivation, approaches, and pros and cons of each category.

\subsection{\bf Contour-based Methods}
In this category, phase contrast microscopy cell population segmentation is accomplished by detecting contours of cells. The underlying assumption is that cells have clear boundaries and the variations within cells are small. 
These rule-based methods can be roughly grouped into two class: Watershed based and Active Contour based. The major problem with these Contour-based algorithms is that contours of cells can be severely corrupted due to shape deformation, artifacts and cell adhesion. Naturally, watershed and active contour methods often produce over-segmentation and under-segmentation, respectively. 

\subsubsection{\bf Watershed}
Beucher et al. proposed a region based segmentation method called Watershed segmentation \cite{Beucher}. The motivation of watershed segmentation came from water flood in natural world that water will fill up basins and dams will prevent water from different basins. The boundaries of objects in images are corresponding to dams in nature and objects segmentation is achieved by delineating boundaries of target objects. Watershed segmentation starts from local minima, therefore, the success of watershed methods is based on the right selection of local minima. Debeir \cite{Debeir} developed a  marked watershed that refines the initial local minima for segmentation. 
The selection of local minima is converted into an classification problem, which uses an assemble principle.
Kachouie et al.~\cite{Watersheddeconvolution} developed a coarse to refine method for watershed segmentation.
The first stage applies a template matching and canny edge detection to find coarse regions of cells. The second stage utilizes watershed transform to refine the obtained result. As appearances of cells are not fixed, template matching is unable to detect cells with events or movement. 
Wang et al. ~\cite{OptimizedWatershed} applied cell peak detection and localization to deal with cell adhesion.
Each cell in a clustered region is defined by the shortest distance between pixels and the detected peaks. However,this works well only when the intensity distribution for each cell varies little, which is not held in real-world due to the presence of artifacts and cell events.


Another common application of watershed is to combine with other segmentation methods. For example, methods in \cite{Tse,LearningtoSegment,Weijun_2} utilize 
the segmentation results by watershed
as the inputs of other methods for further segmentation. We'll introduce these methods in following sections. 



\subsubsection{\bf Active Contour}
Kass et al. \cite{Kass1988} proposed the first active contour model.  Specifically, the best fit of contours is obtained by minimizing an energy function,
also called parametric active contours. While this model suffers from two major problems: sensitive to the initial contours and incapability with concave contours. To overcome these drawbacks, Xu \cite{Xu} proposed the gradient vector flow (GVF) model. Based on GVF, Zimmer et al. further presented a two-step model to segment cells~\cite{Zimmer}. Firstly, the Canny edge map is employed to detect low contrast boundaries. Secondly, a repulsive interaction is proposed to deal with cell adhesion. The major drawback of this method is that the assumption of homogeneous background in images.


In above parametric active contours methods, the parameters need to be computed repeatedly until convergence. To avoid this, Caselles et al.~\cite{Caselles} developed another type of active contours, called geometric active contours. 
Note that the relationship between geometric active contours and parametric active contour is elaborated in \cite{Xu}.
Yang~\cite{Yang2005} introduced a geometric active contours based method, called Narrow Band level set, \cite{Adalsteinsson94afast}. 
The first step is to extract the general position of cells using the gradient information in images.
Next, it employs a fast level set to get a refined segmentation from the previous step.
Tse et al. \cite{Tse} proposed a segmentation method by combining watershed and parametric active contour.
Instead of using a single level set, Vese et al. applied a multiphase level set\cite{Vese2002} to extract more accurate markers. Li \cite{Li2007} et al. presented two-stage level set based segmentation, including morphological pre-segmentation and level set segmentation. The morphological pre-segmentation is based on rolling-ball filter that segments the background and cells. Then, it applies a naive level set to post-process results from the last step.
Gurari et. al.\cite{Gurari} presented a survey on six popular level set methods’ performance on phase contrast microscopy cell data and find the optimal one for each stage of cells’ evolution. 
However, this six stages is not enough to cover all cases in cells' activities, and thus might not suitable for a large dataset.

This geometric active contour is still depend on curve messages, therefore, these methods will fail when testing on a large data set with high complexity.


\subsection{\bf Machine Learning based methods}
In these rule-based methods, PCM cell population segmentation is formulated as a classification problem, which is to learn a statistical model to distinguish cells and non-cell. 
The success of these methods depends on two factors: sufficient training data and effective feature extraction.
In the following, we will present these methods according to their learning styles, including supervised learning, unsupervised learning and semi-supervised and active learning.


\subsubsection{\bf Supervised learning based methods }
Methods in this group need to train models from the data with labels for all pixels.
Pan et al. proposed a cell population segmentation method using Conditional Random Field(CRF) \cite{panHCRF}. 
More formally, interesting pixels are selected from local intensity minima~\cite{Pan_2009_6764} and used as the nodes in CRF graph.  
The nodal potential and edge potential in CRF model are obtained by converting the outputs of soft SVM~\cite{PRML} into posterior class probabilities. 
Another method uses CRF to segment PCM cell population was presented in~\cite{Restif}. The conventional CRF energy function only contains information from a single pixel and pairwise pixels, which can be considered as the first-order and second-order information respectively. 
However, phase contrast microscopy images are often texture-less, therefore, this first-order and second-order information is not enough for representing objects.
Towards this, Kohli et al.~\cite{Kohli} developed a new energy function using $\mathcal{P}^n$ Potts potential function. The energy function is written as below:
\begin{equation}
E(S|I) = \underbrace { \sum_{p\in  I}  E_{1}(S_p|I)}_{first -order} + \underbrace {\sum_{p, q\in  I}  E_{2}(S_p,S_q|I)}_{second-order}+\underbrace {\sum_{C\in  I}  E_{C}(S_C|I) }_{higher-order},
\end{equation}
where $S_p$ is a pixel and $S_q$ is its neighbor; $C$ a clique in the image.
The $\mathcal{P}^n$ Potts potential is a higher order clique potential of arbitrary clique size. 
He et al. proposed a two-stage method combining machine learning and shape information in cells.
Cells are segmented via a learning model and further refined by curve evolution~\cite{Weijun_1}. 
Here, it defines features of cells as intensity distribution, gradient magnitude Distribution, and Earth Mover’s Distance. 
Wang et al. ~\cite{Weijun_2} integrate AdaBoost with SVM to perform PCM cell population segmentation. This method contains three tasks: cell detection, foreground segmentation, and individual cell segmentation. An AdaBoost classifier is trained by using Mexican Hat wavelet features for detecting cell centers. 
The detection result is used to as the seed of watershed segmentation for the segmentation of whole cells. 
Experimental results show that methods in ~\cite{Weijun_2,Weijun_1} fail to segment clustered cells due to watershed method has genetic problem of dealing with cell adhesion.

Funke et al.~\cite{LearningtoSegment} introduced hierarchical method for PCM cell population segmentation problem. Firstly, a Random Forest classifier~\cite{ilastik} is applied to obtain the boundaries of cells. This followed by a watershed segmentation to get rough cells.
Secondly, it applies a structured SVM classifier~\cite{Tsochantaridis} to learned a cost function for extracting accurate boundary of cells.
This method needs neither manually feature engineer nor parameter tuning, which reduces human interference.


Li et al.~\cite{Li2007,kang2008} solved the PCM cell segmentation problem using a Bayesian classifier.
It extracts color histograms as features for training a model.
Yin et al. ~\cite{YinBayesian} also addressed this problem using a Bayesian classifier.
The feature of pixels is computed based on intensity histogram in a neighbouring window, and is compressed by clustering to reduce redundancy.


\subsubsection{\bf Unsupervised learning based methods }
Since labeling pixels is often expensive, 
unsupervised learning style is another option to deal with PCM cell population segmentation.
Mualla et al.~\cite{Mualla2014} presented 
a two-step clustering scheme for this problem. 
The first step is to detect cells using the scale invariant feature transform (SIFT) feature~\cite{Mualla2}. To separate cells from the background, it employs a two class K-medians clustering, where it uses  difference of gaussian (DOG) and  local variance as features.
The final cell segmentation is also implemented based on clustering, in particular, the number of clusters is obtained by a self-labeling algorithm~\cite{Pan_2009_6764}.


Zhang et al.~\cite{Zhang2014} developed a segmentation method based on Random Forest and correlation clustering. Similar to \cite{LearningtoSegment}, 
Random Forest classifier~\cite{ilastik} is used to identify the boundary of cells.
Then, it constructs superpixels  based on the resulting boundary probability and obtains an adjacency graph by weighting superpixels.
The final cell segmentation is converted into a graph partitioning problem. 

Instead of using natrual image features as in ~\cite{Mualla2014,Zhang2014},
Su et al.~\cite{spectralSu} introduced a new feature by taking advantages of PCM imaging principles, called phase retardation feature. 
The key is to partition images by clustering pixels, where it computes pairwise similarities using phase retardation features. Note that this feature depend on research on methods in section 2.3. 

\subsubsection{\bf Semi-supervised and Active learning}

Both Semi-supervised and Active learning need a limit number of labels for training, therefore, this is a merit
in dealing with the PCM cell segmentation problem due to the limit access to labels.


Su et al.~\cite{semi-supervised} first proposed a method that uses semi-supervised learning architecture to segment cells in PCM images . It learns a dictionary using the proposed retardation features, which will be elaborated in Section 2.3. 
Next, each pixel is represented by a coefficient vector using the obtained dictionary, where these pixels have labels.
This is followed by a label propagation to infer cells in the unlabeled data over an affinity matrix that characterizes the relationship between unlabeled atoms and labeled atoms,
which is to optimize the following objective:
\begin{equation}
    f({\bf Y}_u) = trace\left ( [{\bf Y}_l;{\bf Y}_u]^T {\bf L} \begin{bmatrix}
{\bf Y}_l\\ 
{\bf Y}_u
\end{bmatrix} \right )
\end{equation}
where ${\bf L}$ is a Laplacian matrix; ${\bf Y}_l$ and ${\bf Y}_u$ are the matrices of labeled and unlabeled atoms, respectively.
The final segmentation result ${\bf Y}_{u}^{*}$ is computed as:  
\begin{equation}
    {\bf Y}_{u}^{*} = {\bf L}_{uu}^{-1} {\bf W}_{ul}{\bf Y}_{l}
\end{equation}
where ${\bf W}_{ul}$ represent similarity between unlabeled and labeled atoms.
In order to lower human intervention in this method,  Su et al. \cite{Hangsu2} developed a method that combines semi-supervised learning and active learning. 
They first detect errors from the results of semi-supervised segmentation, and then correct them by a human.
Next, the corrected information will be propagated over the whole graph to fix other similar errors. These two steps will be alternatively proceed until all errors are eliminated. 
To further minimize human intervention, Su et al.~\cite{Hangsu3} improved both the error detection criterion and information propagation strategy in~ \cite{Hangsu2}. 
Different to the affinity matrix updating strategy in~\cite{Hangsu2} that propagates information to whole graph in each iteration, Su et al. developed an faster solution using augmented graph \cite{Zhu03combiningactive}, in which only a part of the matrix will be updated in each iteration. 
Although experimental results show excellent performance in PCM cell population segmentation, one potential weakness of active learning based methods is that the error collection must be comprehensive, otherwise, these methods will fail when the errors in the testing are different from those in the training data.




\subsection{\bf Restoration based methods }

\begin{figure}
\centering\includegraphics[width=0.9\linewidth]{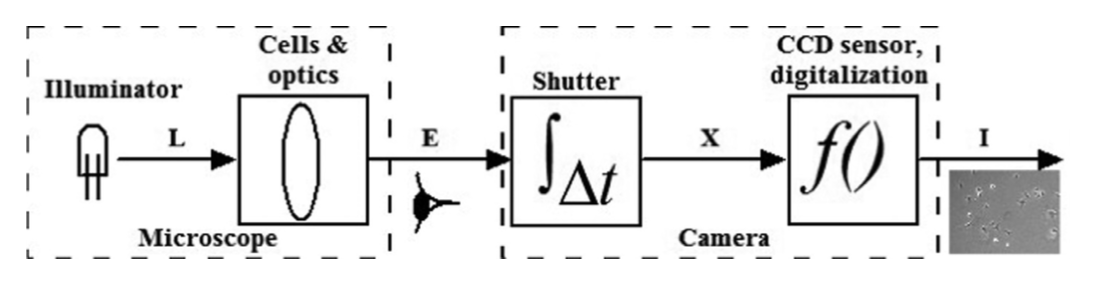}
\caption{\label{Fig:Phase contrast microscopy imaging model} Phase contrast microscopy imaging model}
\end{figure}

In previous categories,  all methods treat PCM images as natural images and employ methods from  natural images' analysis.
However, this neglects the difference between 
PCM images and natural images due to their significant different imaging principles.
In this section, we focus on methods of dealing PCM image cell population segmentation by analyzing its underlying structures.
The phase contrast microscopy imaging system is composed of Microscope and Camera devices, as shown in  Figure~\ref{Fig:Phase contrast microscopy imaging model} \cite{cellsensitive}. Under this system, the light (L) passes through cells and optics, then produces the irradiance (E) for observation by eyes or cameras. 
Recently, several methods have been proposed to model this imaging process and convert cell segmentation as a restoration problem.
For example, 
methods in \cite{Restoring,understanding,Zhaozheng,semi-supervised} formulate the imaging procedure as a linear model. 
Besides,  methods ~\cite{cellsensitive,cellsensitive2} have been trying to estimate the relationship between exposure duration and intensities with a cell-sensitive camera response function from the camera part. 
In both methods, cells are restored and cell segmentation can be implemented by simple thresholding.  

\subsubsection{\bf Preconditioning}
In this approach, researchers restore artifact-free microscopy images by approximating PCM imaging system as a regularized quadratic model. 
Li et al.~\cite{likang} first demonstrated that Differential Interference Contrast (DIC) microscopes imaging system can be approximated by a linear imaging model. 
Inspired by this work, Yin et al.~\cite{Restoring,understanding} proposed a preconditioning algorithm, where cells in PCM images can be segmented from a restored artifact-free image by thresholding.
More formally, this model is written as as follows: 
\begin{equation}
\mathbf{g} \approx \mathbf{Hf}
\end{equation}
where $\mathbf{H}$ is the convolution kernel using the point spread function (PSF); $\mathbf{g}$ is the  observed image and $\mathbf{f}$ is the artifact-free image. 
This is solved by an iterative algorithm~\cite{Sha}.

In \cite{Restoring,understanding}, the underlying assumption is that the phase retardation caused by cells is tiny, almost zero. However, this assumption does not hold in practical due to the presence of bright cells like mitosis and apoptosis cells. To fix this, Su etc. \cite{Zhaozheng} improved this model by generalizing the phase retardation $\theta (x)$ as a linear combination as below:
\begin{equation}
e^{i\theta (x)}\approx \sum_{m=0}^{M-1}\varphi _{m}(x)e^{i\theta (m)},   s.t.\varphi _{m}(x)\geq 0
\end{equation}
where $\theta _{m} = 0,\frac{2 \pi}{M},...,\frac{2m \pi}{M},...\frac{2(M-1)\pi}{M}$,   $\varphi _{m}(x)$ is the coefficient. Then the restoration problem can be converted to the following optimization problem:
\begin{equation}
\displaystyle{\min \sum_{k=0}^{N-1}\left \{ \left \| \Psi _{mk} \right \|_{1}     +\omega_s\ \Psi _{mk}^{T}\mathbf{ L} \Psi _{mk}  \right \}, \begin{Bmatrix}
\left \|  g- \sum_{k=0}^{N-1}  \mathbf{H}_{mk}  \Psi _{mk}     \right \|_2 <\varepsilon \\ 
\Psi _{mk}^{T}\geq 0 
\end{Bmatrix}
}
\end{equation}
where $\bf L$ is a Laplacian matrix to describe similarity between spatial pixel neighbors; $\mathbf{\Psi}$ is the vectorized coefficient matrix.
Each pixel in the images is represented by a feature vector $\left \{ \Psi _{m1},\Psi _{m2},....\Psi _{mk} \right \}$. 
The segmentation is accomplished by K-means clustering on the obtained feature vectors.

\subsubsection{\bf Cell-sensitive phase contrast microscopy Imaging}

\begin{figure}
\centering\includegraphics[width=0.9\linewidth]{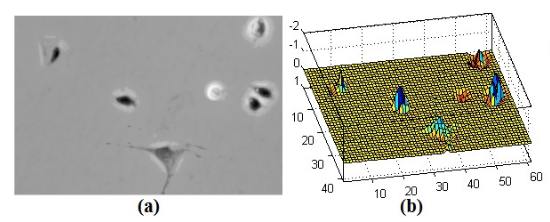}
\caption{\label{Fig:cell sensitive segmentation}cell sensitive segmentation, (a)Restored irradiance image; (b) Irradiance map}
\end{figure}

In contrast to preconditioning methods that focus on microscope imaging part, 
Zhao et atl.~\cite{cellsensitive,cellsensitive2} developed methods by taking advantage of camera information in the PCM system.
The motivation is that different exposure duration $\Delta t$ of the camera will change the image intensity $I$ as below
\begin{equation}
\label{eq:csf}
I =f(E \Delta t) 
\end{equation}
where $f$ is the camera response function. 
To estimate $f$ , the following optimization problem is formulated:
\begin{equation}
\begin{aligned}
&O(g,E) = \sum_{i=0}^{N}\sum_{j=0}^{P}\left \{  g(I_{ij})-\log E_i -\log\Delta t_j  \right \}^2\\
&+\alpha \sum_{i\in [1,N]}(\log E_i)^2 +\beta \sum_{I}[\omega (I)g^{''}(I)]^2
\end{aligned}
\end{equation}
where $E_i$ is the irradiance at the $i_{th}$ pixel; $I_{ij}$ is the intensity at the $i_{th}$ pixel with exposure duration $\Delta t_j$. By taking the derivatives of $g$ and $E$ and equating to zero, this problem will be converted to an overdetermined problem, which can be addressed by a singular value decomposition method.
Once the camera response function $f$ is estimated, the irradiance $E_i$ can be obtained based on its pixel values:
\begin{equation}
    \log E_i = \frac{\sum_{j=1}^{P}\omega (I_{ij})(g(I_{ij})-\Delta t_j)    }{\sum_{j=1}^{P}\omega (I_{ij})}
\end{equation}
A high contrast map between cells and background is then obtained, as shown in Figure~\ref{Fig:cell sensitive segmentation}. Therefore, cell segmentation is accomplished by simply thresholding. 
Later on,
another segmentation method was developed by analyzing camera exposure~\cite{cell-sensitive3}, which computes
 Maximally Stable Extremal Regions(MSER) \cite{matas2004robust} under each exposure settings.
It gets a binary map of cells by accumulating these obtained MMSERs using the frequency of pixels in each MMSER. 
The final segmentation is accomplished by applying a Graph-cut on MMSER. 







\subsection{\bf Discussion}

We reviewed PCM cell population segmentation methods in three major categories. Note that some methods can be assigned into multiple categories. For example, some unsupervised learning methods~\cite{Zhang2014,spectralSu} employ ideas from restoration based methods. 

The Contour-based methods heavily rely on boundaries information of cells. However, such information in PCM images is unreliable because of the shape deformation and artifacts. Therefore, these rule-based methods have an obvious limit on PCM cell population segmentation. 

For machine learning methods, these aim to find statistical models of cells. The key to the success is extracting representative features. While traditional hand-crafted features have limited power since PCM cell images are textureless.
It is promising to integrate deep learning for feature extraction. 
Furthermore, machine learning based methods depend on training datasets and labels.
With only a small amount of labeled data are available, semi-supervised learning, active learning and unsupervised learning methods have more potentials than supervised learning  methods.

Restoration based methods are the most effective method, however, these methods are often having high computation costs, which is not suitable for a large-scale datasets. Due to the high complexity, fast solutions of these methods are needed to reduce costs.

\section{Database and Performance Evaluation}

Although many PCM cell population segmentation methods have been developed many years, only a few methods have experimented on large-scale datasets, which contain a large number of images and each image has a high density of cells. 
Moreover, researchers evaluate their methods on different datasets using different evaluation measures.
These raise concerns on the performance evaluation and comparison.
In this section, we aim to give a review on the characteristics of public phase contrast microscopy cell image datasets
 and  commonly used evaluation metrics in this topic. 

    
    

\begin{figure}[htbp]
\centering
\includegraphics[width=2.2in]{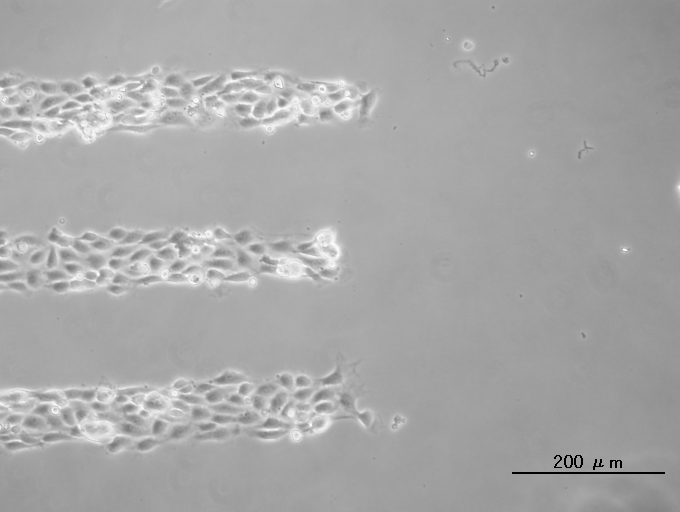}
\includegraphics[width=2.2in]{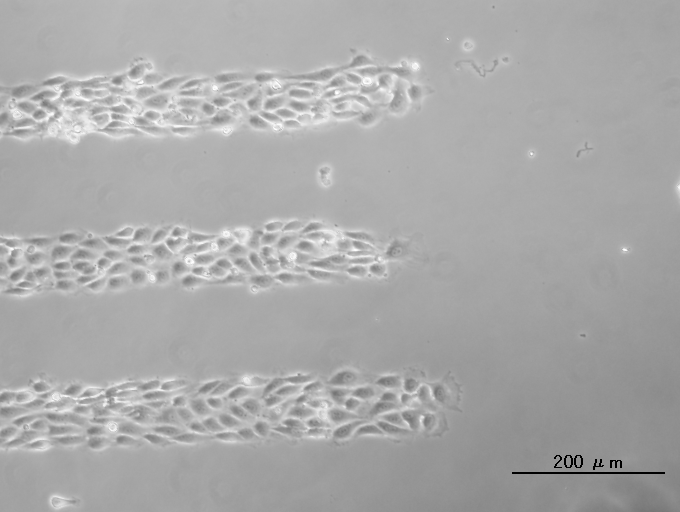}
\caption{\label{Fig:JYdata}Sample images from \cite{panHCRF}: bovine aortic endothelial cells}
\end{figure}

\begin{figure}[htbp]
\centering
\includegraphics[width=2.2in]{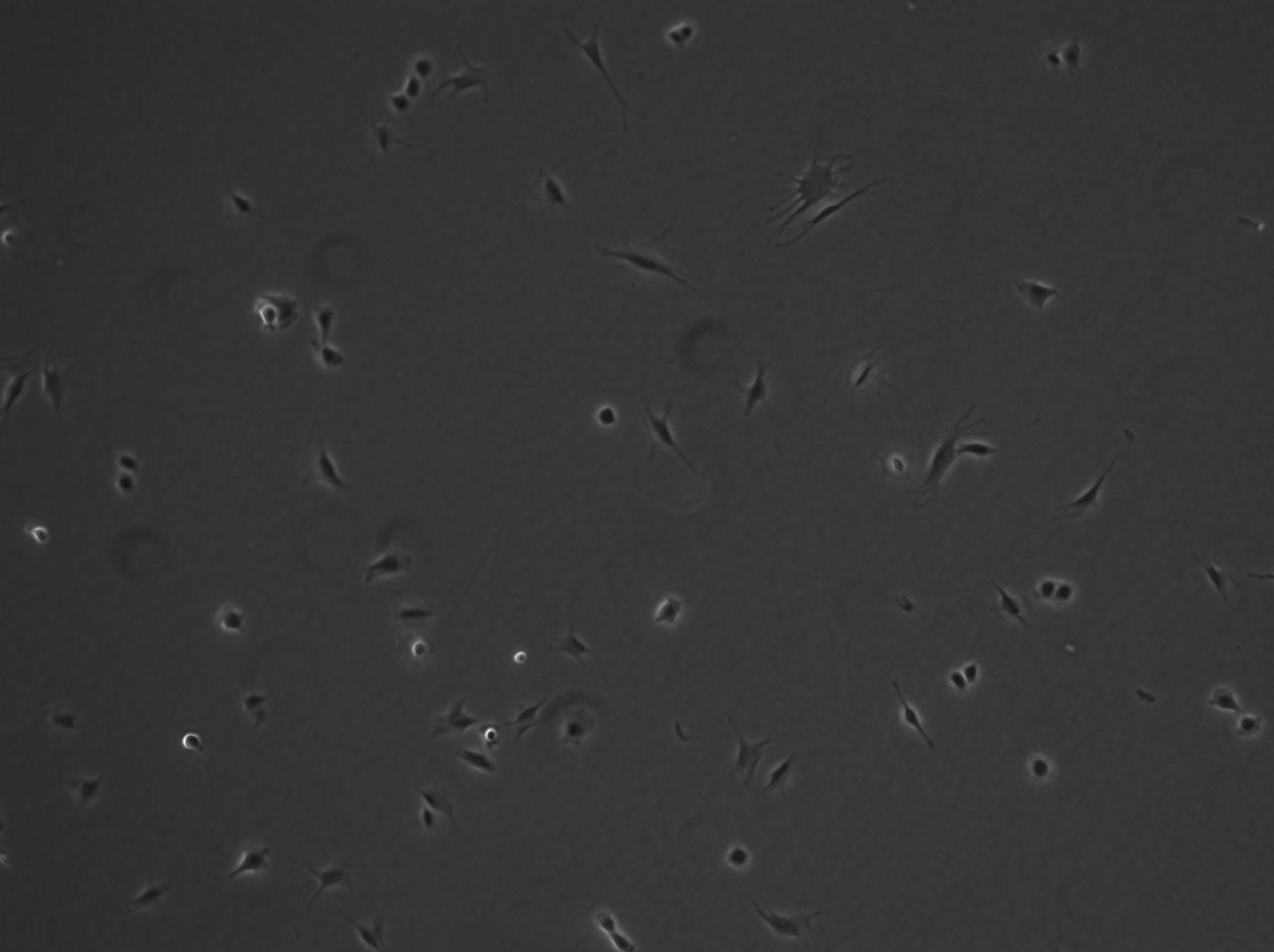}
\includegraphics[width=2.2in]{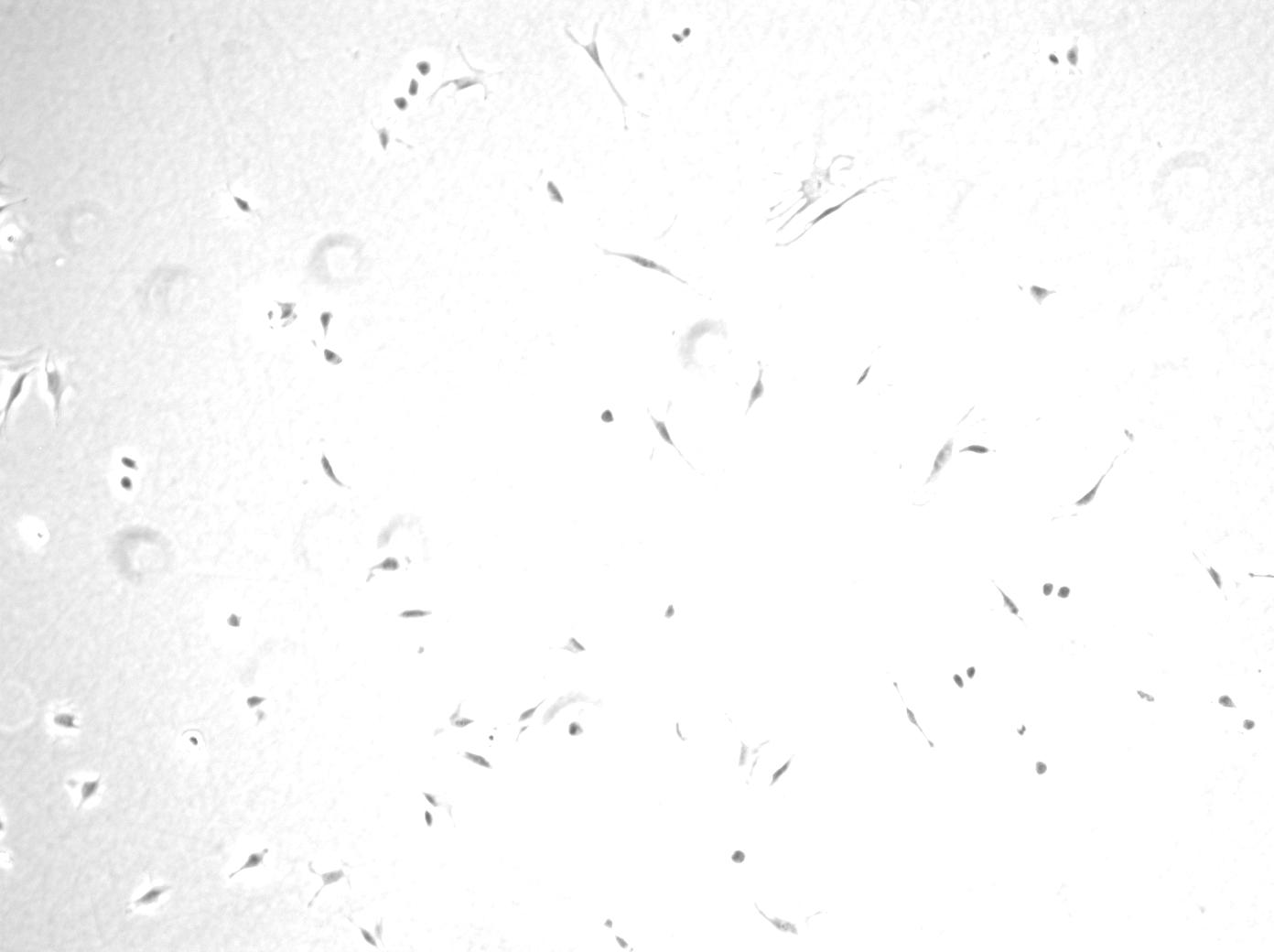}
\caption{\label{Fig:YINSENSI}Sample images used in \cite{cellsensitive,cellsensitive2}: with exposure of 100ms, 500ms}
\end{figure}

\begin{figure}[htbp]
\centering
\includegraphics[width=2.2in]{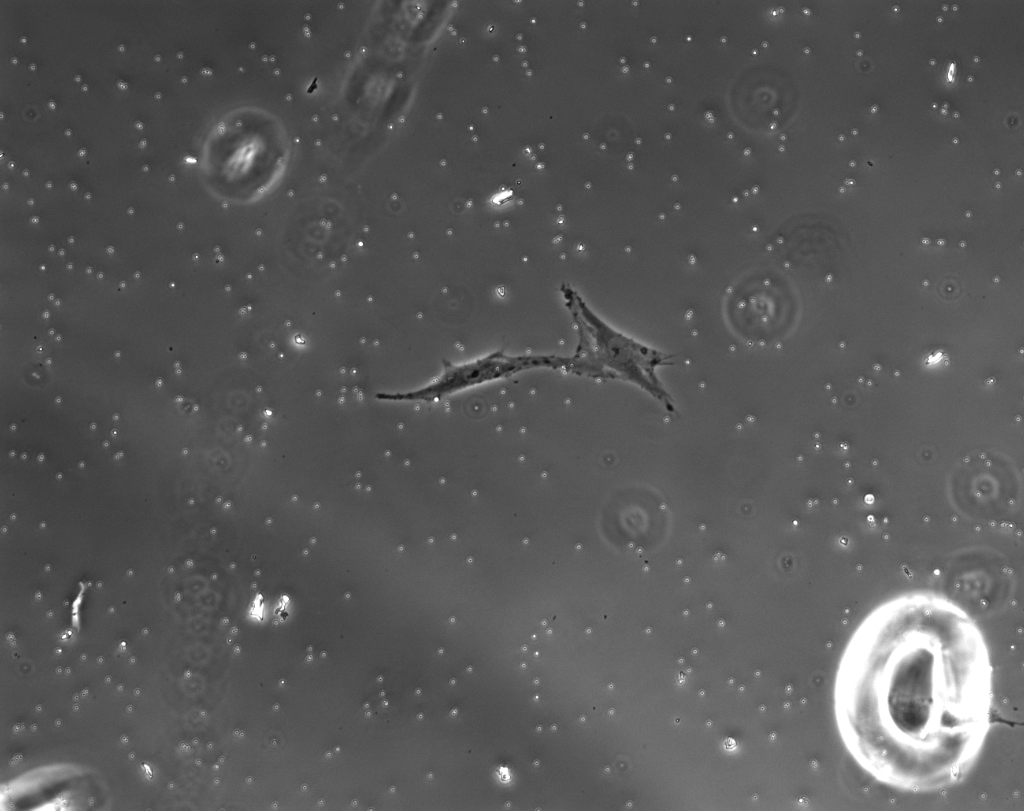}
\includegraphics[width=2.2in]{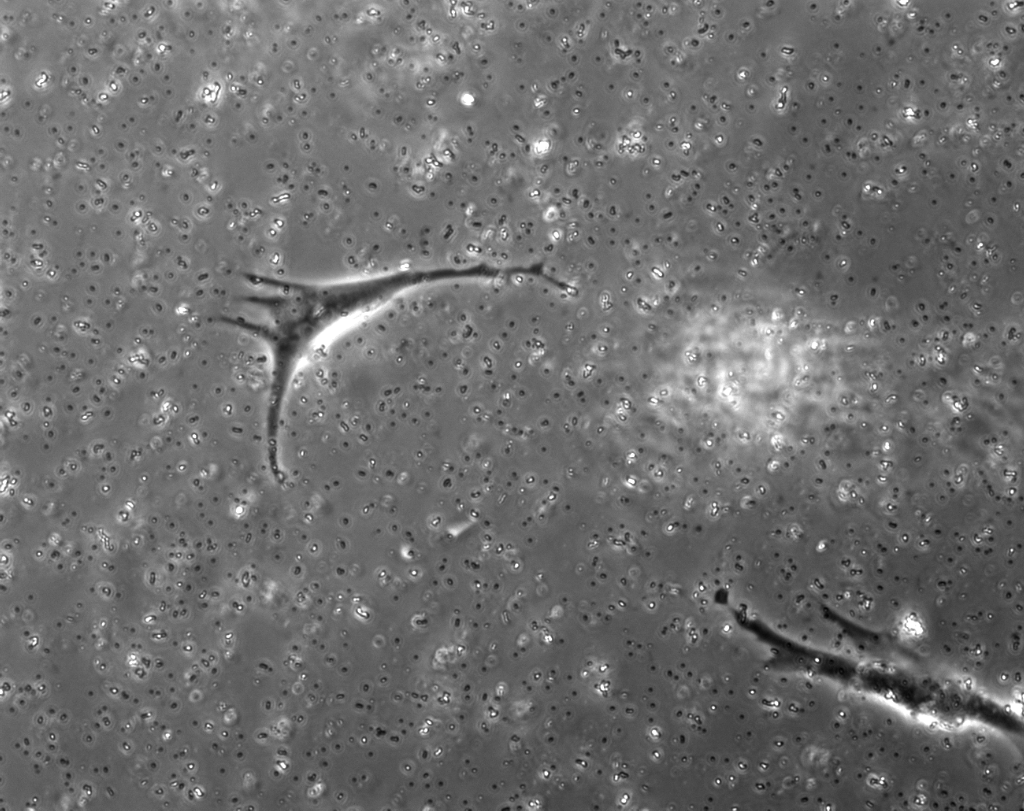}
\caption{\label{Fig:BUDATA}Sample images used in \cite{Gurari}}
\end{figure}

\subsection{ \bf Phase  Contrast  Microscopy  cell Image Datasets}
\begin{figure}[htbp]
\centering
\includegraphics[width=2.2in]{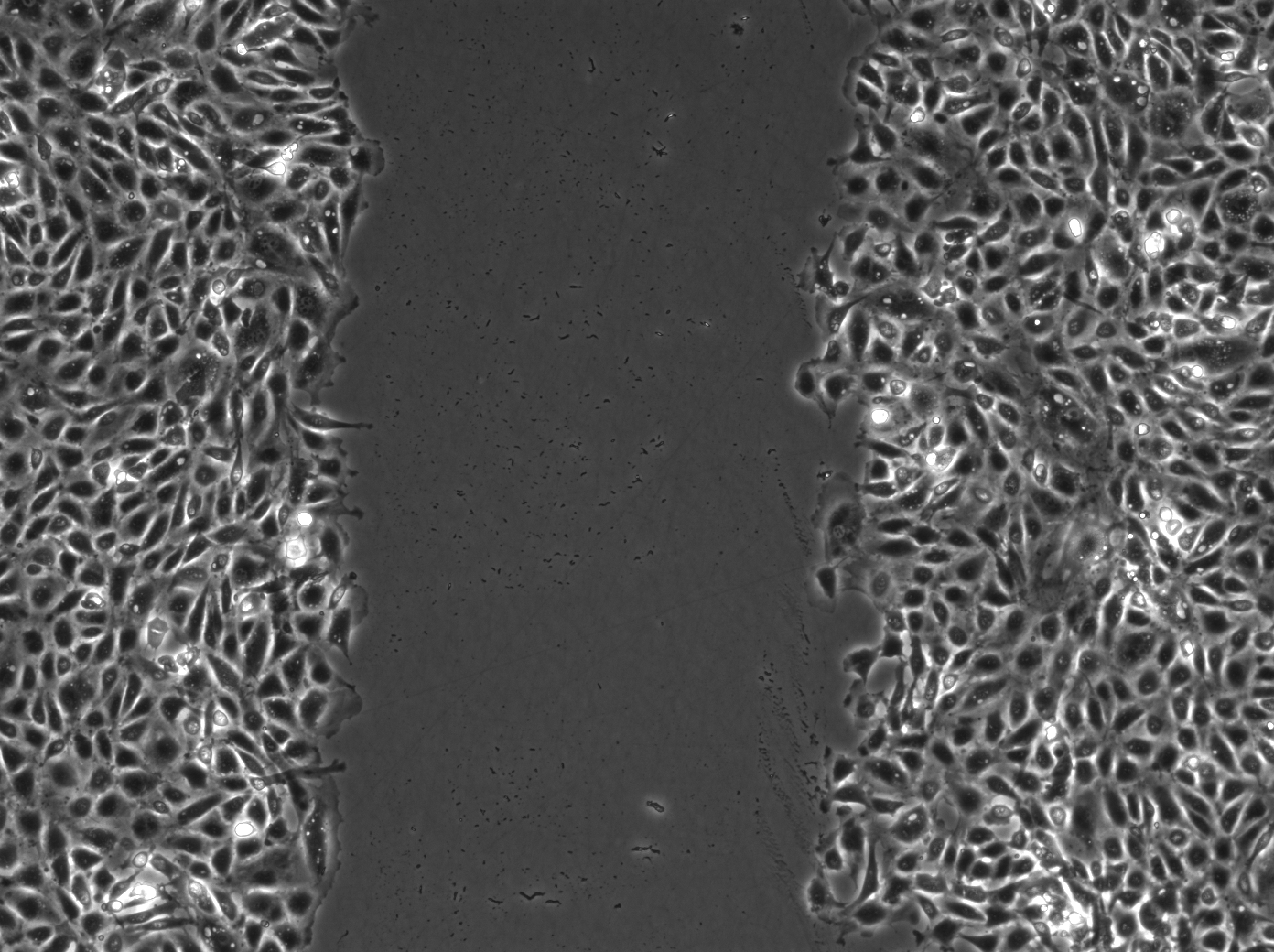}
\includegraphics[width=2.2in]{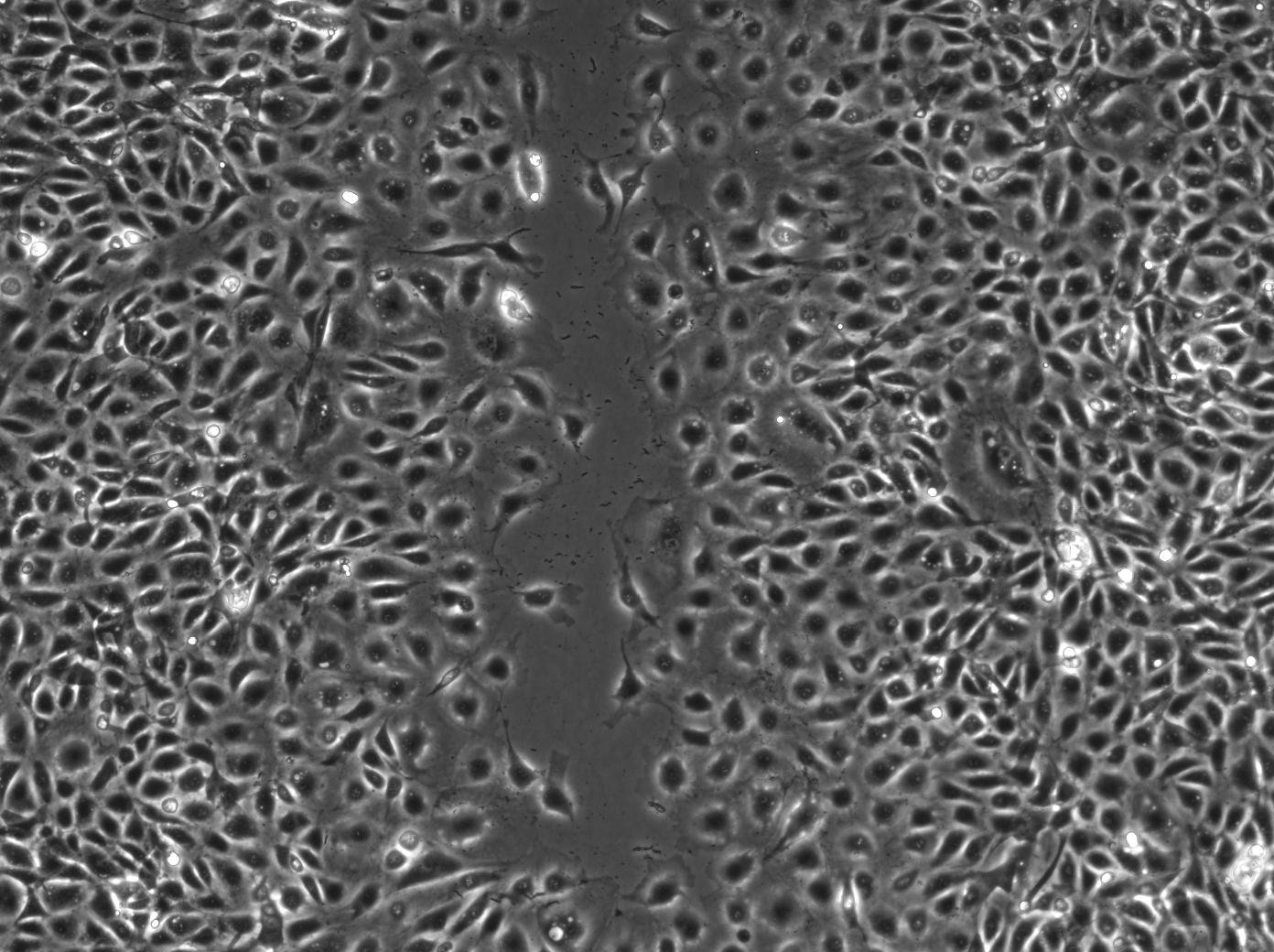}
\includegraphics[width=2.2in]{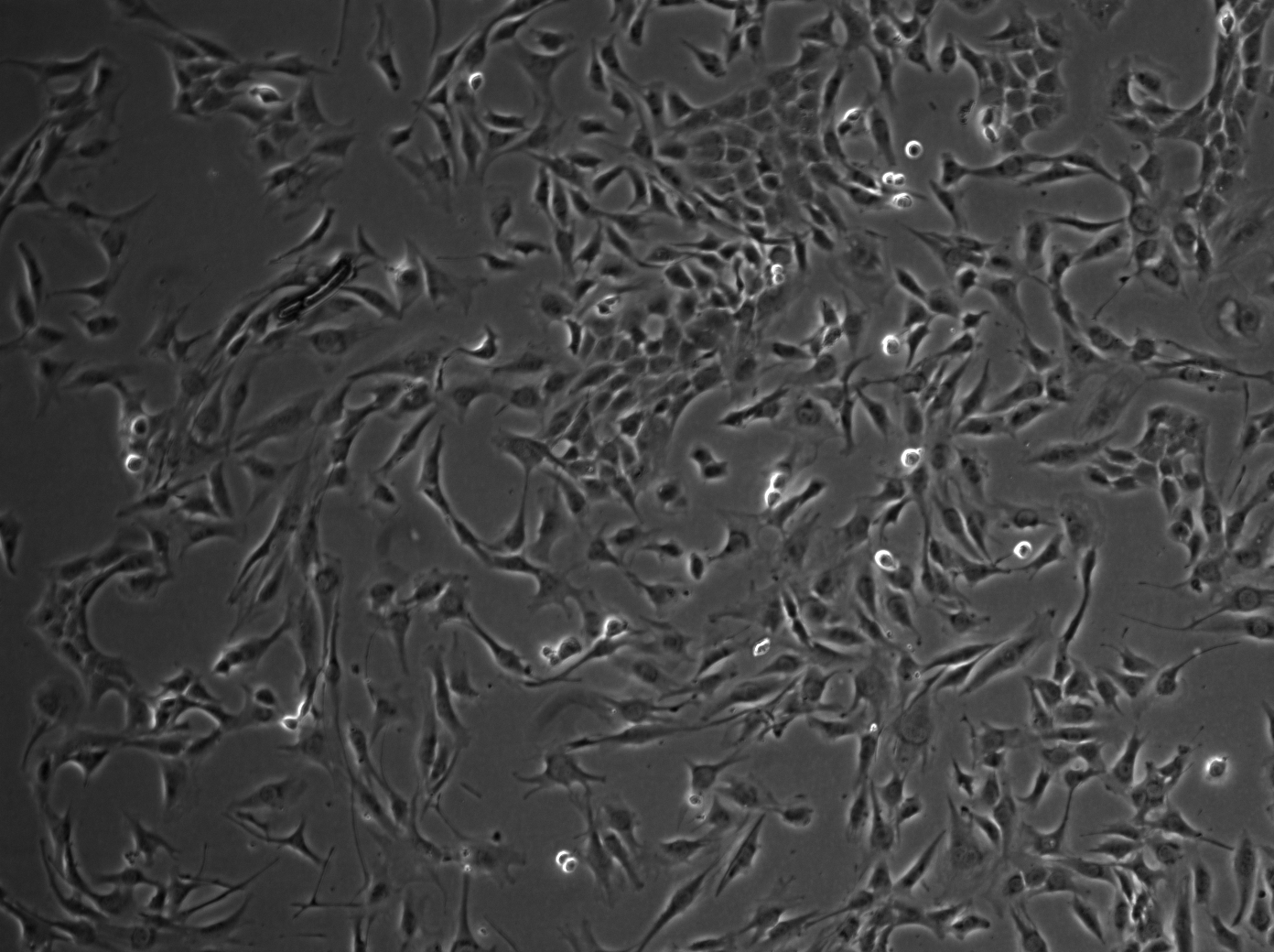}
\includegraphics[width=2.2in]{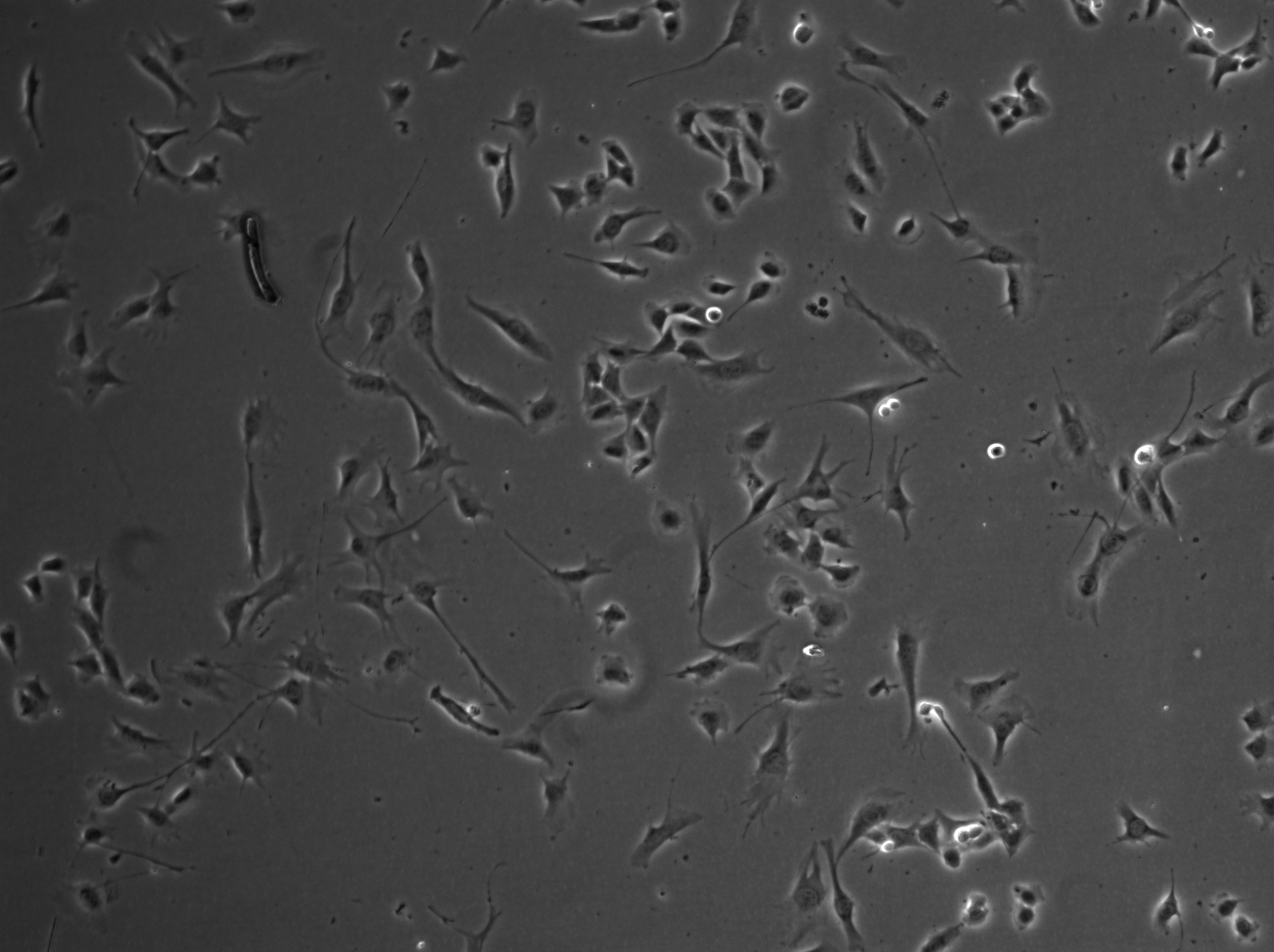}
\caption{\label{Fig:databaseYin}Sample images used in \cite{semi-supervised,understanding,Zhaozheng,Hangsu3,Hangsu2}}
\end{figure}

The database applied in \cite{panHCRF} consists of  two different types of cells: bovine aortic endothelial cells and C2C12 muscle stem cells ,which is available at \url{http://www.albany.edu/celltracking/downloads.html}. Each type of cell has 10 training images and 10 testing images. The average number of cells in bovine aortic endothelial cells images and C2C12 muscle stem cells images are 4,900 and 9,800, respectively. We show two samples from this data as in Fig. ~\ref{Fig:JYdata}.
The database used in \cite{semi-supervised,understanding,Zhaozheng,Hangsu3,Hangsu2} contains three different kind of cells images, which are bovine aortic endothelial cells, C2C12 stem cells and muscle stem cells, which is available at \url{http://www.albany.edu/celltracking/downloads.html}.
The detail specifications are presented in Table ~\ref{table:dataYin}, and some sample images are shown in Fig.~\ref{Fig:databaseYin}.
Li et al. created a dataset \cite{Li2007,kang2008} contains two image sequences of MG-63 osteosarcoma cells with 150 frames and another two sequences of proprietary amnion epithelial stem cells with 256 frames. 
In \cite{cellsensitive,cellsensitive2}, cell images are obtained in six different exposure durations( [50 100 200 300 400 500]ms) for three different cell dishes, as shown in Fig.~\ref{Fig:YINSENSI}. 
The cell database from BU-BIL \cite{Gurari} contains 235 images in total and 152 are phase contrast images, which 35 rat smooth muscles cells, 70 rabbit smooth muscle cells, and 47 fibroblasts, which is available at \url{ http://www.cs.bu.edu/∼betke/Biomedical ImageSegmentation}.
We show some samples in 
Fig. ~\ref{Fig:BUDATA}. 
 
\begin{table}[ht]
\footnotesize
\caption{Database specifications }
\centering
\small
\resizebox{\textwidth}{!}{%
\begin{tabular}{c c c c c}
\hline\hline
 & Image Number & Cell number per image & Image size & Type of cell  \\ [0.5ex] 
\hline
Sequence 1 &589 &500$\sim$800+ & 1040x1392 &bovine aortic endothelial cell \\
Sequence 2 &632 &50$\sim$300+ & 696x520& muscle stem cell\\
Sequence 3 &383 &300+  & 1040x1392& C2C12 myoblastic stem cell\\

\hline
\end{tabular}
\label{table:dataYin}
}
\end{table}

\subsection{\bf Performance Evaluation}

Most papers of phase contrast microscopy cell population segmentation are evaluated on different databases rather than a few public benchmark datasets. 
In fact, only a few of them are tested a common database,
where we show the comparison of them in Table ~\ref{table:comparison}.
So far, the commonly used metric in this topic are accuracy, precision, recall, Normalized Probabilistic Rand index (NPR)~\cite{NPR}, Tannimoto coefficient (TC). These can be roughly grouped into two major categories: pixel-level and cell-level metric, where NPR belongs to cell-level measure and the others are pixel-level measures. 
The successful cell segmentation should not only count the number of segmented pixels but also measure whether or not these pixels belong to corresponding cells. Therefore, both cell-level and pixel-level measures should be considered in the evaluation criterion. 



\begin{table}
\footnotesize
\centering
\caption{Segmentation results on the same database (Acc: Accuracy, Re: Recall; Pre: Precision)}
\label{table:comparison}
\resizebox{\textwidth}{!}{%
\begin{tabular}{l|l|l|l|l|l|l|l|l|l|l|l|l|l|l|l|l|l|l|l|l|l||l|l|l||l|l|}
\hline
\multirow{2}{*}{\bf Methods} &\multicolumn{6}{c|}{ \bf Sequence 1} & \multicolumn{6}{c|}{\bf Sequence 2} & \multicolumn{6}{c|}{\bf Sequence 3} \\ 
\cline{2-19}  & Acc  & TC & NPR & F & Re & Pre &  Acc &  TC&NPR & F& Re & Pre & Acc  & TC & NPR & F & Re & Pre \\ 
\cline{1-19}Precoditioning \cite{understanding} & 0.97     &  0.83  &  0.75   & 0.8676  &  0.858   &  0.878   &  0.9      &0.31         &  0.27     &   0.1749   &0.157 & 0.348  &    &  &  & 0.3212   & 0.347 & 0.314 \\ 

\cline{1-19}Generalized precoditioning. \cite{Zhaozheng}   &      &    &     & 0.9319  &  0.921   &  0.943   &        &       &      &   0.9319 & 0.942 & 0.967 &    &  &  &    0.9412& 0.932 &  0.963\\

\cline{1-19}Semi-supervised learning\cite{semi-supervised}  &      &  0.94  &0.88     &   &   0.963  & 0.893    &        &     0.92   & 0.92      &    & 0.979 &0.863    &    &  &  &    &  &   \\ 
\cline{1-19}Bayesian classifier\cite{YinBayesian}&      &    &     &  0.93 &0.9     &0.967     &        &        &       &   0.911  & 0.925&0.898  &    &  &  &    0.893&  0.854& 0.936 \\
\cline{1-19} Clustering \cite{spectralSu}  &      &    &   0.87  &   &     &     &        &        &  0.85     &   & &   &    &  &  &    &  &  \\ 
\cline{1-19}Active learning 1\cite{Hangsu2}  &      &    &  0.95   &   &     &     &        &        &    0.95   &   & &   &    & & 0.95  &    &  &  \\ 
\cline{1-19}Bounded Active learning\cite{Hangsu3}  & 0.96     &    &     &   &     &     &  0.96      &        &       &   & &   &   0.96 &  &  &    &  &  \\ 
\hline
\end{tabular}
}
\end{table}

\section{CONCLUSION}
This paper aims to present a thorough study of phase contrast microscopy cell population segmentation, including methodologies, datasets and metrics. 
We present the motivation, general steps, and pros and cons for methods in this area.
We also highlight two important open issues in this problem, which are the missing of benchmark datasets and inconsistent evaluation metrics. 
We hope researchers in this community can build a benchmark dataset and a unified metric
for better comparison.







\end{document}